\documentclass[letterpaper]{article} 
\usepackage{xcolor}   
\usepackage{amssymb}  
\usepackage[submission]{aaai2026}  
\usepackage{booktabs}
\usepackage{times}  
\usepackage{helvet}  
\usepackage{courier}  
\usepackage[hyphens]{url}  
\usepackage{graphicx} 
\usepackage{amsmath}
\DeclareMathOperator{\clip}{clip}
\usepackage{natbib}  
\usepackage{caption} 
\usepackage{algorithm}
\usepackage{algorithmic}
\usepackage{pifont}
\usepackage{newfloat}
\usepackage{listings}
\DeclareCaptionStyle{ruled}{labelfont=normalfont,labelsep=colon,strut=off} 
\floatstyle{ruled}
\newfloat{listing}{tb}{lst}{}
\floatname{listing}{Listing}
\title{ChronoForge-RL: Chronological Forging through Reinforcement Learning for Enhanced Video Understanding}

\makeatletter
\gdef\showauthors@on{T}
\renewcommand{\@author}{Kehua Chen}
\long\gdef\affiliations#1{\gdef\affiliations_{\if T\showauthors@on#1\fi}}
\affiliations{%
  Independent Researcher%
}
\makeatother

\usepackage{bibentry}

\begin{document}

\maketitle

\begin{abstract}

Current state-of-the-art video understanding methods typically struggle with two critical challenges: (1) the computational infeasibility of processing every frame in dense video content and (2) the difficulty in identifying semantically significant frames through naive uniform sampling strategies. In this paper, we propose a novel video understanding framework, called ChronoForge-RL, which combines Temporal Apex Distillation (TAD) and KeyFrame-aware Group Relative Policy Optimization (KF-GRPO) to tackle these issues. Concretely, we introduce a differentiable keyframe selection mechanism that systematically identifies semantic inflection points through a three-stage process to enhance computational efficiency while preserving temporal information. Then, two particular modules are proposed to enable effective temporal reasoning: Firstly, TAD leverages variation scoring, inflection detection, and prioritized distillation to select the most informative frames. Secondly, we introduce KF-GRPO which implements a contrastive learning paradigm with a saliency-enhanced reward mechanism that explicitly incentivizes models to leverage both frame content and temporal relationships. Finally, our proposed ChronoForge-RL achieves 69.1\% on VideoMME and 52.7\% on LVBench compared to baseline methods, clearly surpassing previous approaches while enabling our 7B parameter model to achieve performance comparable to 72B parameter alternatives, a 10× improvement in performance-to-parameter ratio.

\end{abstract}


\section{Introduction} 
The proliferation of Large Language Models (LLMs) \cite{chatgpt} and Multimodal Large Language Models (MLLMs) \cite{bai2025qwen2.5vl} have marked a new era in open-world understanding. A natural and critical frontier for these models is the extension from static images to the dynamic and complex realm of videos. Videos encapsulate a rich, continuous spatiotemporal stream of information, from subtle facial expressions to large-scale event dynamics. Consequently, the ability to accurately and efficiently interpret this content, identifying key events, capturing semantic cues, and performing deep causal reasoning, remains a fundamental challenge in artificial intelligence research\cite{fu2025video,wang2024lvbench}. This task demands that models not only integrate and reason over complex details across temporal dimensions but also discern the nuanced semantic relationships and causal chains that bind them.

Recent advances in LLMs and MLLMs, powered by their robust contextual modeling and large-scale pre-training, have significantly enhanced multimodal comprehension\cite{bai2025qwen2.5vl,zhang2025deep}. Notably, the dramatic expansion of context windows to over one million tokens \cite{team2024gemini,yang2025qwen2} has opened unprecedented avenues for video analysis. However, despite these strides, a critical bottleneck persists. The dense information volume of videos, characterized by important semantic events interspersed with visually similar but less informative frames, poses significant challenges for current models' processing capabilities.

\begin{figure}
    \centering
    \includegraphics[width=1.0\linewidth]{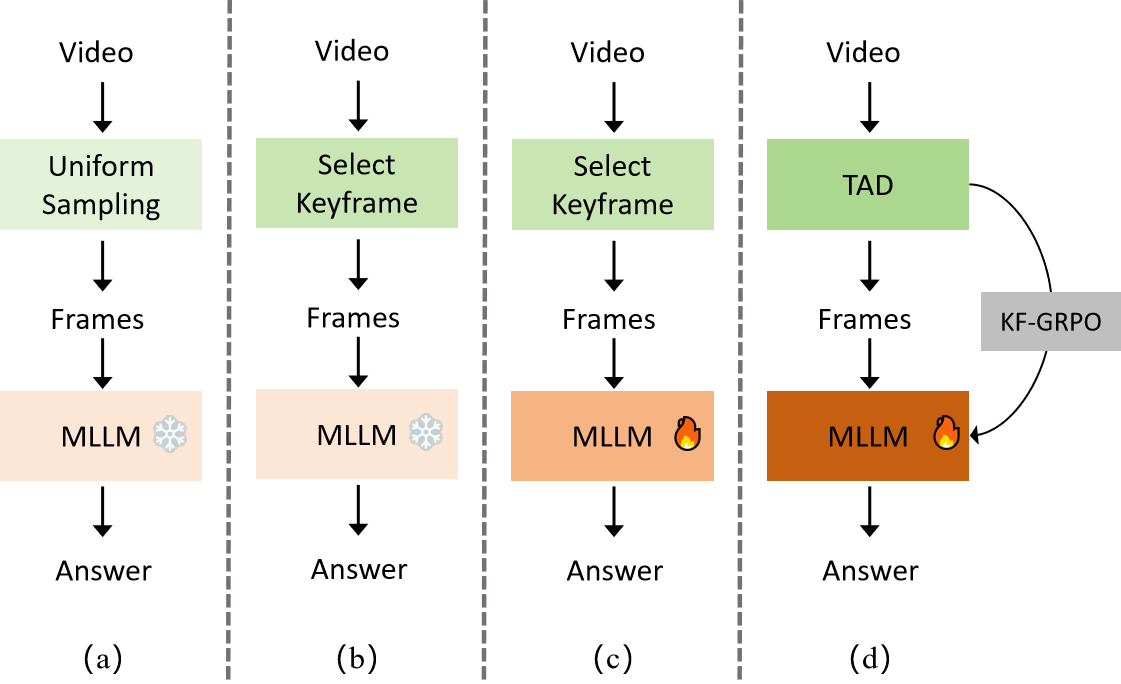}
    \caption{A comparison between typical MLLM video understanding frameworks. (a) Basic uniform sampling pipeline. (b) Keyframe selection approaches. (c) Model Fine-Tuning Based Spatiotemporal Adaptive Compression. (d) Our proposed ChronoForge-RL integrating TAD and KF-GRPO.}
    \label{fig2}
\end{figure}

Figure \ref{fig2} presents a comprehensive overview of video understanding frameworks for MLLMs. Typically, as shown in Figure 1(a), the basic MLLM video understanding process involves processing the entire sequence of video frames, often using uniform sampling. This approach has a critical drawback: it frequently overlooks decisive moments, leading to distorted or even erroneous interpretations of the video narrative. This highlights the urgent need for intelligent and efficient keyframe selection mechanisms, which, however, still require substantial computational resources. Recent methods \cite{wang2025adaretake} have evolved towards keyframe selection approaches, as illustrated in Figure 1(b). This mechanism bridges the gap between the powerful semantic understanding of MLLMs and the unique spatiotemporal characteristics of dynamic videos, creating an analysis framework that preserves key information while reducing computational complexity. Specifically, videos are simplified by selecting only representative frames, allowing MLLMs to process these sparse keyframes instead of the full continuous video sequence. The overall effectiveness of keyframe-based methods critically depends on the quality of the selected frames. To bridge this gap, recent keyframe selection methods \cite{tang2025adaptive, wang2025adaretake, li2024videochat} aim to identify a small, representative subset of frames based on criteria like user query relevance or visual redundancy. While effective at reducing computational load, these approaches share fundamental limitations: (1) their reliance on a sparse set of frames is often insufficient for tasks requiring fine-grained temporal analysis or long-horizon reasoning; and (2) they scale poorly, as increasing the number of selected frames to capture more detail leads to prohibitive computational overhead.

Alternatively, some methods address this problem from a multimodal large language model fine-tuning perspective by employing specialized adaptation techniques that effectively align video understanding capabilities with language comprehension, as shown in Figure 1(c). Such methods \cite{shen2024longvu,yang2025pvc} focus on carefully adjusting pre-trained model parameters to better handle the unique challenges of video content, enabling comprehensive processing of video sequences while maintaining semantic coherence and temporal relationships in the final representations.

Differing from existing works, our approach introduces Temporal Apex Distillation (TAD) for keyframe selection and systematically integrates this selection philosophy into the entire video understanding training pipeline, enabling the model to maintain temporal coherence while capturing semantic inflection points and significantly improving performance on complex video reasoning tasks. To address these challenges, we propose a novel framework built upon two key components. First, we design a highly adaptive keyframe extraction scheme that identifies frames of maximal semantic and temporal importance. Second, and more critically, we present KeyFrame-aware Group Relative Policy Optimization (KF-GRPO), an extension of T-GRPO \cite{feng2025video} that incorporates keyframe selection directly into a reinforcement learning loop. KF-GRPO explicitly encourages robust temporal reasoning based on the selected keyframes by training the model with two sequences: one containing correctly ordered keyframes and a “negative” sequence with shuffled or non-key frames inserted. A positive reward is given only if the model’s response accuracy on the pristine keyframe sequence exceeds that on the negative sample, thereby compelling the model to learn not only the content of keyframes but also the intrinsic value of their correct temporal ordering.

Our main contributions are summarized as follows:

\begin{itemize} 
\item We introduce Temporal Apex Distillation (TAD), a differentiable keyframe selection algorithm that systematically identifies semantic inflection points through variation scoring, inflection detection, and prioritized distillation, addressing the fundamental trade-off between temporal information retention and computational efficiency in video understanding. 
\item We develop KeyFrame-aware Group Relative Policy Optimization (KF-GRPO), the first approach that combines keyframe selection with GRPO reinforcement learning, integrating keyframe awareness with contrastive temporal reasoning through a novel saliency-enhanced reward mechanism that explicitly incentivizes models to leverage both frame content and temporal relationships. 
\item We demonstrate state-of-the-art performance among open-source models on benchmark datasets, achieving 69.1\% on VideoMME and 52.7\% on LVBench, enabling our 7B parameter model to achieve performance comparable to 72B parameter alternatives, a 10× improvement in performance-to-parameter ratio that makes advanced video understanding viable for resource-constrained applications. \end{itemize}

\section{Related Work}

\subsection{Multimodal Large Language Models for Video}
Recent advances in video understanding can be broadly categorized into training-based and training-free paradigms. In the training-free paradigm, Adaretake\cite{wang2025adaretake} proposes an adaptive allocation strategy to optimize resource distribution for video processing. For temporal allocation, it partitions the video into segments and assigns compression ratios based on the similarity between adjacent frames. For layer-wise allocation, it dynamically adjusts compression ratios across different layers according to attention scores derived from video prompts. AKS\cite{tang2025adaptive-aks} introduces plug-and-play modules for keyframe selection and pre-filtering of multi-modal information in video models. QuoTA\cite{luo2025quota} presents a pre-training-free module that performs query-guided frame-level importance assessment for visual token allocation. The query-guided token selection is critical as it aligns visual processing with task-specific requirements, enabling efficient utilization of token budgets while preserving semantically relevant content. In the training-based paradigm, NVILA\cite{liu2024nvila} enhances its model architecture through two-stage improvements: first, by expanding spatial and temporal resolutions, followed by compressing visual tokens. This "scale-then-compress" approach enables NVILA to efficiently handle high-resolution images and videos. Qwen2.5-VL\cite{bai2025qwen2.5vl} achieves efficient and precise understanding of arbitrary-sized images and videos through dynamic resolution processing, absolute time-aligned MRoPE, and efficient visual encoding. It further leverages a 4.1 trillion token high-quality dataset to comprehensively enhance multi-modal capabilities.

\subsection{Multimodal Large Language Model Reasoning}
Recent work has reignited interest in enhancing the reasoning capabilities of LLM, particularly in enabling them to effectively tackle complex, multi-step tasks. The emergence of DeepSeek-R1\cite{guo2025deepseek} has been especially impactful, demonstrating that rule-based reinforcement learning frameworks can significantly stimulate the latent reasoning abilities of LLMs, even when driven solely by outcome-based rewards. Notably, DeepSeek-R1 achieves robust chain-of-thought reasoning without the need for intermediate supervision, highlighting the power of well-designed reward structures and strategic policy optimization in fostering self-improving models. Building on these foundational insights, several follow-up studies leverage the advantage of GRPO to tackle multimodal tasks. One notable advancement, Video-R1\cite{feng2025video}, introduces T-GRPO, wherein the model simultaneously processes temporally coherent frame sequences and temporally shuffled counterparts. A reward signal is crafted to favor correct temporal ordering; if the model performs better on ordered sequences, it receives positive reinforcement. This mechanism incentivizes temporal reasoning without requiring dense annotation or stepwise supervision. TW-GRPO\cite{dang2025reinforcing} advances the reward shaping and policy learning components by integrating information-theoretic weighting and soft reward mechanisms inspired by Intersection-over-Union (IoU) metrics from video localization tasks. These innovative approaches further demonstrate that nuanced, weakly-supervised signals can unlock strong reasoning capabilities in multimodal LLMs. However, despite promising advancements, a critical and underexplored challenge remains: scaling reasoning to long-form videos, where selecting semantically rich keyframes plays a pivotal role in task performance. Efficiently identifying and leveraging such keyframes in lengthy, information-dense video streams is non-trivial, and represents a compelling next step in pushing the frontiers of LLM-powered video understanding.

\section{Method}
\subsection{Overview}

\begin{figure*}[t]
    \centering
    \includegraphics[width=1.0\linewidth]{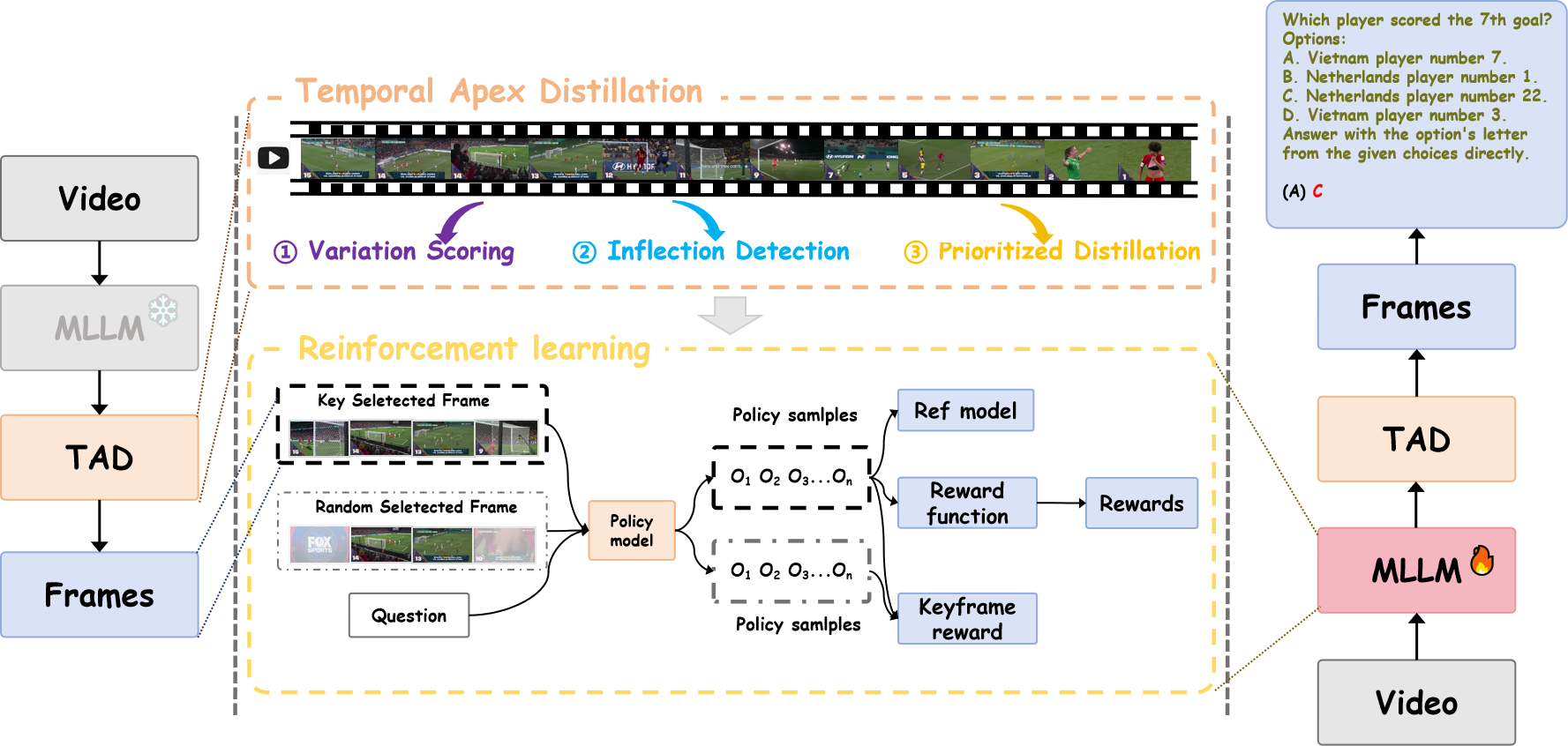}
    \caption{Overview of the ChronoForge-RL Framework. The pipeline starts with video input, where keyframe selection leverages an MLLM integrated with Temporal Apex Distillation (TAD). After offline data processing, extracted keyframes and non-keyframes feed into the KF-GRPO algorithm to generate completions and compute rewards. A multi-level soft reward mechanism integrates keyframe ratios and spatiotemporal sequences for nuanced feedback, guiding model updates. This enables prioritization of informative frames, enhancing overall performance.}
    \label{fig:enter-label}
\end{figure*}

We present a framework for video understanding that combines intelligent keyframe selection with temporal-aware policy optimization. Our approach consists of two key components: TAD and KF-GRPO. TAD introduces a differentiable keyframe selection mechanism that identifies semantically significant frames through a three-stage process: variation scoring to quantify temporal changes, inflection detection to locate critical turning points, and prioritized distillation to select the most informative frames. Building upon this foundation, KF-GRPO extends traditional policy optimization by incorporating keyframe awareness and temporal reasoning. It employs a novel contrastive learning paradigm that leverages both sequential and disordered frame arrangements, combined with a saliency-enhanced reward mechanism that prioritizes temporal understanding. The synergy between these components creates a comprehensive framework that not only reduces computational overhead by focusing on essential frames but also enhances temporal reasoning capabilities through structured policy learning. Our method advances the state-of-the-art by seamlessly integrating differentiable frame selection with temporal-aware policy optimization, offering a robust solution for complex video understanding tasks while maintaining computational efficiency.

\subsection{Temporal Apex Distillation (TAD)}

\begin{algorithm}[tb]
\caption{Adaptive Temporal Feature Distillation}
\label{alg:temporal_distillation}
\textbf{Input}: Feature sequence $F \in \mathbb{R}^{T \times N \times C}$, Budget $K$, Window $W$\\
\textbf{Output}: Distilled sequence $F'\in \mathbb{R}^{K \times N \times C}$
\begin{algorithmic}[1]
\STATE // Compute temporal variation
\FOR{$t=1$ to $T-1$}
    \STATE $V[t] \leftarrow 1 - \text{CosineSimilarity}(F[t-1], F[t])$
\ENDFOR
\STATE $V \leftarrow \text{Mean}_{\text{spatial}}(V)$ \quad // Aggregate across spatial dim
\STATE // Identify and amplify inflection points
\STATE $P \leftarrow \text{FindLocalMaxima}(V, W)$
\STATE $S \leftarrow V$; $S[P] \leftarrow S[P] + \omega$
\STATE // Select top-K frames
\STATE $I \leftarrow \text{argTopK}(S, K)$
\STATE $F' \leftarrow F[I]$
\STATE \textbf{return} $F'$
\end{algorithmic}
\end{algorithm}

To efficiently process videos, we propose Temporal Apex Distillation (TAD), detailed in Algorithm \ref{alg:temporal_distillation}. This method is designed to select the most informative $K$ keyframes from an input sequence of $T$ visual features, $F \in \mathbb{R}^{T \times N \times C}$, where $N$ is the number of spatial patches or tokens. Rather than relying on uniform sampling, TAD identifies frames capturing significant semantic changes. Crucially, our implementation supports two distinct modes of operation: a synchronous (frame-level) mode where a frame is either kept or discarded entirely, and an asynchronous (patch-level) mode where key-patches are selected independently across the temporal dimension. For clarity, our main description focuses on the more intuitive synchronous mode. The selection process is structured into three sequential steps: Variation Scoring, Inflection Detection, and Prioritized Distillation.

\textbf{Variation Scoring.} 
The first step quantifies the degree of content change across the timeline. We measure the importance of each frame based on its feature deviation from the preceding frame. For each timestep $t$, we calculate the temporal variation $V_t$ by computing the dissimilarity between adjacent frame features $F_{t-1}$ and $F_t$. This is performed for all $N$ spatial locations, resulting in a raw dissimilarity matrix of shape $T \times N$. For the synchronous mode, we aggregate these dissimilarities across the spatial dimension $N$ using max pooling to obtain a robust, global measure of change for each frame. This yields a single variation score sequence $V \in \mathbb{R}^T$, where higher values indicate more significant temporal changes.

\textbf{Inflection Detection} 
While the variation score $V$ highlights high-activity frames, we hypothesize that inflection points, moments where the rate of visual change peaks, are more critical for understanding video narratives. To locate these critical moments, we perform local maxima detection on the variation scores $V$. Instead of a naive iterative search, we employ a highly efficient and differentiable approach leveraging a 1D max-pooling operation with indices. A frame at index $t$ is identified as a local maximum if its own index matches the index of the maximum value within its sliding window of size $W$. This provides a vectorized, GPU-friendly implementation for finding the set of inflection point indices $P$. These indices represent the pivotal turning points in the video's narrative.

\textbf{Prioritized Distillation} 
The final step integrates the global variation scores with the detected inflection points to select the final keyframes. We construct a definitive selection score $S$, initialized with $V$. Crucially, we prioritize the identified inflection points $P$ by adding a large, constant boosting weight, $\omega$ (\textup{see Appendix 1.1 for more details)}, to their corresponding scores in $S$. This mechanism ensures that frames capturing critical turning points are almost always selected, provided they are within the budget $K$. We then determine the final set of keyframe indices $I$ by performing an argTopK operation on the prioritized scores $S$. To preserve the chronological flow of the video, these selected indices $I$ are subsequently sorted in ascending order.

Finally, we extract the distilled feature sequence $F' \in \mathbb{R}^{K \times N \times C}$ by gathering features from the original sequence $F$ using the indices $I$. Importantly, this gathering operation is differentiable with respect to the input features $F$. This allows gradients from the subsequent reinforcement learning task to flow back to the visual encoder, implicitly optimizing the encoder to produce features that enhance the effectiveness of this selection strategy.

\subsection{KF-GRPO}

Our KeyFrame-aware Group Relative Policy Optimization (KF-GRPO) framework employs contrastive learning between sequentially arranged and disordered frame inputs while incorporating keyframe extraction to mitigate inefficiencies from redundant or noisy frames in extended videos. Keyframes represent the core informative elements, and by prioritizing them, KF-GRPO refines the policy's focus, enhancing both training speed and reasoning precision.

\textbf{Training Frame Curation.} For a video sequence $V$ comprising $T$ frames, we employ our TAD algorithm to derive a reduced set $S$ of $T_s = \lfloor \delta T \rfloor$ salient frames, with $\delta \in (0, 1]$ denoting the selection fraction. This yields: 
\[ 
S = \text{TAD}(V, \delta),
\] 
where $S = {s_1, s_2, \ldots, s_{T_s}}$. These keyframes are pre-processed for the entire training dataset and integrated via image loading, enabling targeted training that distinguishes between essential and auxiliary frames to bolster temporal comprehension.

\textbf{Paired Response Generation.} Given a query $q$ linked to the video, we form two response ensembles based on differential frame selections: 
\begin{itemize} 
\item 
\emph{Sequential keyframes:} $S$ containing all extracted keyframes preserved in their original chronological sequence. 
\item \emph{Hybrid disordered frames:} $\hat{S}$ comprising approximately 50\% keyframes and 50\% non-keyframes, with the entire sequence randomly shuffled to disrupt temporal coherence. \textup{(see Appendix 1.2 for more details)}. 
\end{itemize}

\textbf{Saliency-Enhanced Reward.}
Denote $c$ and $\hat{c}$ as the accuracy rates for the sequential and disordered ensembles, respectively. The saliency-boosted temporal incentive $r_s$ is formulated as:
\[
r_s = \mathbb{I}[c > \hat{c}],
\]
where $\mathbb{I}[\cdot]$ is the indicator function yielding 1 if true and 0 otherwise.

The aggregated reward for each output $a_j$ is formulated through a component architecture: 
\[
Q_j = b_j + r_s \cdot \mathcal{A}(a_j) 
\] where $b_j$ constitutes the primary reward signal that quantifies response quality through multiple dimensions: semantic accuracy with respect to ground truth, logical and structural coherence of reasoning chains, and distributional alignment with expert demonstrations or high-capacity reference models. The saliency-weighted accuracy term $r_s \cdot \mathcal{A}(a_j)$ serves as a targeted reinforcement component that explicitly rewards temporal understanding when chronologically ordered keyframes yield superior performance over temporally disrupted sequences. This theoretically grounded reward formulation creates a learning signal that simultaneously optimizes for content comprehension and temporal reasoning, compelling the policy to develop sophisticated representations that capture both the semantic richness of individual frames and the causal relationships that connect them across the temporal dimension.

\textbf{Relative Advantage Estimation.}
$R_i$ denotes the raw reward of the $i$-th response, while $R_j$ represents its normalized and shifted advantage estimate used to guide policy updates. We assess the scaled advantage for each output in its ensemble:
\[
R_j = R_i - \mu_E + \frac{\sigma_E}{2},
\]
where $\mu_E$ and $\sigma_E$ are the mean and variance of the ensemble's rewards, respectively, providing a shifted normalization that highlights superior performances in keyframe-driven contexts.

\begin{multline*}
J_{\text{KF-GRPO}}(\theta) = \mathbb{E}_{q, \{a_j\}} \Bigg[ \frac{1}{M} \sum_{j=1}^M \\
\clip\left( \frac{\pi_\theta(a_j \mid q, S)}{\pi_{\theta_{\text{init}}}(a_j \mid q, S)}, 1-\eta, 1+\eta \right) \mathcal{A}_{j} 
- \gamma \text{KL}(\pi_\theta \| \pi_{\text{base}}) \Bigg]
\end{multline*}

where the clipping bounds policy shifts, and the KL term ensures proximity to a reference policy $\pi_{\text{base}}$. This formulation embeds saliency-aware incentives into the optimization process, fostering stable and effective updates centered on pivotal frames. $\pi_{\text{init}}$ restricts policy updates via ratio clipping, while $\pi_{\text{base}}$ provides a stable reference through KL regularization.

\begin{table*}[ht]
\begin{center}
\resizebox{\textwidth}{!}{
\begin{tabular}{lccccccc}
\toprule
Model & Params & Trainable & LVBench &\multicolumn{4}{c}{VideoMME} \\
 &&&Val&Long&Medium&Short&Overall\\
\midrule
\textit{Proprietary Models}\\
\addlinespace[0.1cm]
GPT4-V&-&\textcolor{green}{\ding{51}}&-&53.5&-&-&59.9 \\
GPT4-o&-&\textcolor{green}{\ding{51}}&27.0&65.3&-&-&71.9 \\
Gemini-1.5-Pro&-&\textcolor{green}{\ding{51}}&33.1&67.4&-&-&75.0 \\
\midrule
\textit{Open-Sourse MLLMs}\\
\addlinespace[0.1cm]
VideoLLaVA\cite{lin2023videollava}&7B&\textcolor{green}{\ding{51}}&-&36.2&38.0&45.3&39.9 \\
VITA-1.5\cite{fu2025vita1.5}&7B&\textcolor{green}{\ding{51}}&-&47.1&54.2&67.0&56.1 \\
mPLUG-Owl3\cite{ye2024mplug}&7B&\textcolor{green}{\ding{51}}&43.5&50.1&57.7&70.0&59.3 \\
VideoLLaMA3\cite{zhang2025videollama}&7B&\textcolor{green}{\ding{51}}&45.3&54.9&63.7&80.1&66.2 \\
Oryx-1.5\cite{liu2024oryx}&7B&\textcolor{green}{\ding{51}}&30.4&-&-&-&58.3 \\
QuoTA\cite{luo2025quota}&7B&\textcolor{red}{\ding{55}}&-&55.7&64.9&77.1&65.9\\
Video-XL\cite{shu2024videoxl}&7B&\textcolor{green}{\ding{51}}&-&-&-&-&55.5 \\
ViLaMP\cite{cheng2025scaling-vilamp}&7B&\textcolor{green}{\ding{51}}&-&-&-&-&67.5 \\
AdaReTaKe\cite{wang2025adaretake}&7B&\textcolor{red}{\ding{55}}&51.2&58.3&-&-&67.7\\
AKS\cite{tang2025adaptive-aks}&7B&\textcolor{red}{\ding{55}}&-&-&-&-&65.3\\
NVILA\cite{liu2024nvila}&8B&\textcolor{green}{\ding{51}}&-&54.8&62.2&75.7&64.2 \\
ByteVideoLLM\cite{wang2024dynamic-bytevideollm}&14B&\textcolor{green}{\ding{51}}&-&56.4&62.9&74.4&64.6 \\
Qwen2-VL\cite{wang2024qwen2}&7B&\textcolor{green}{\ding{51}}&42.4&53.8&-&-&63.3 \\
Qwen2.5-VL\cite{bai2025qwen2.5vl}&7B&\textcolor{green}{\ding{51}}&45.3&55.6&-&-&65.4 \\
LLaVA-Video\cite{zhang2024video}&72B&\textcolor{green}{\ding{51}}&-&61.5&-&-&70.6 \\
Qwen2.5-VL\cite{bai2025qwen2.5vl}&72B&\textcolor{green}{\ding{51}}&47.3&63.9&-&-&72.6 \\

\midrule
\textit{Reinforcement learning-based model}\\
\addlinespace[0.1cm]
VRAG-RL\cite{wang2025vrag}&7B&\textcolor{green}{\ding{51}}&43.4&54.6&63.8&76.7&65.1 \\
Video-R1\cite{feng2025video}&7B&\textcolor{green}{\ding{51}}&44.7&54.4&63.7&75.5&64.6 \\

TW-GRPO\cite{dang2025reinforcing}&7B&\textcolor{green}{\ding{51}}&43.7&55.7&64.3&76.5&65.5 \\
MiMo-VL\cite{team2025kimi} & 7B & \textcolor{green}{\ding{51}} & 45.9 & 55.2 & 67.7 & 76.5 & 66.3 \\
\textbf{ChronoForge-RL} &\textbf{7B}&\textcolor{green}{\ding{51}}&\textbf{52.7}&\textbf{60.0}&\textbf{67.0}&\textbf{80.3}&\textbf{69.1} \\
\bottomrule
\end{tabular}}
\caption{Video understanding accuracy(\%) on LVBench and VideoMME. The methods are divided into three categories: Proprietary Models, Open-Source MLLMs, and Reinforcement Learning-based models. ChronoForge-RL based on Qwen2.5-VL achieved state-of-the-art results on the Reinforcement Learning-based models and Open-Source MLLM benchmarks among models with comparable parameter scales.}
\vspace{-1em}
\label{table1}
\end{center}
\end{table*}

\section{Experiments}
\subsection{Experimental Setups}

\begin{table*}[ht]
\begin{center}
\resizebox{\textwidth}{!}{
\begin{tabular}{lccccccc}
\toprule
Model & Params & use TAD & LVBench & \multicolumn{4}{c}{VideoMME} \\
&&& Val & Long & Medium & Short & Overall \\
\midrule
VRAG-RL\cite{wang2025vrag} & 7B & \textcolor{red}{\ding{55}} & 43.4 & 54.6 & 63.8 & 76.7 & 65.1 \\
VRAG-RL\cite{wang2025vrag} & 7B & \textcolor{green}{\ding{51}} & 50.1 & 55.4 & 62.8 & 76.6 & 65.0 \\
Video-R1\cite{feng2025video} & 7B & \textcolor{red}{\ding{55}} & 44.7 & 54.4 & 63.7 & 75.5 & 64.6 \\
Video-R1\cite{feng2025video} & 7B & \textcolor{green}{\ding{51}}  & 49.7 & 57.0 & 63.6 & 75.3 & 65.3 \\
TW-GRPO\cite{dang2025reinforcing} & 7B & \textcolor{red}{\ding{55}} & 43.7 & 55.7 & 64.3 & 76.5 & 65.5 \\
TW-GRPO\cite{dang2025reinforcing} & 7B & \textcolor{green}{\ding{51}}  & 50.3 & 59.5 & 65.8 & 77.5 & 67.6 \\
MiMo-VL\cite{team2025kimi} & 7B & \textcolor{red}{\ding{55}} & 45.9 & 55.2 & 67.7 & 76.5 & 66.3 \\
MiMo-VL\cite{team2025kimi} & 7B & \textcolor{green}{\ding{51}}  & 43.7 & 48.0 & 59.2 & 75.4 & 60.8 \\
RL-Baseline & 7B & \textcolor{red}{\ding{55}} & 44.8 & 58.2 & 64.8 & 75.7 & 66.2 \\
ChronoForge RL & 7B & \textcolor{green}{\ding{51}}  & \textbf{52.7(+7.9)} & \textbf{60.0(+1.8)} & \textbf{67.0(+2.2)} & \textbf{80.3(+4.6)} & \textbf{69.1(+2.9)} \\
\bottomrule
\end{tabular}}
\caption{Performance(\%) comparison of TAD applied to various RL-based video understanding models. We evaluate the improvements TAD brings to baseline models on LVBench and VideoMME.}
\vspace{-1em}
\label{table3}
\end{center}
\end{table*}

\textbf{VideoMME} \cite{fu2025video} is a benchmark dataset specifically designed for evaluating multi-modal video understanding systems. The dataset consists of 900 videos with a total duration of 256 hours, spanning 30 diverse domains. It contains 2,700 manually annotated complex multiple-choice question-answer pairs, enabling comprehensive evaluation of models' capabilities in understanding and reasoning about multi-modal video content. To facilitate fine-grained assessment across temporal scales, VideoMME is partitioned into three subsets based on video duration: short-form (less than 2 minutes), medium-form (4-15 minutes), and long-form (30-60 minutes), with each subset containing 900 question-answer pairs. This stratified structure enables systematic analysis of model performance across different temporal contexts.

\noindent\textbf{LVBench} \cite{wang2024lvbench} represents one of the most comprehensive and challenging benchmarks for long-form video understanding. The dataset comprises 103 hour-long videos accompanied by 1,549 annotated multiple-choice questions. These test samples span multiple cognitive dimensions, including entity recognition, event understanding, key information extraction, temporal localization, and higher-order reasoning. By focusing on hour-long temporal reasoning and in-depth comprehension assessment, LVBench not only presents significant challenges for existing video understanding models but also serves as a crucial benchmark for evaluating long-context video analysis capabilities.

\subsection{Implementation details}
\textbf{Training Details.} We train our ChronoForge-RL model based on Qwen2.5-VL-7B using 8 NVIDIA A100 GPUs. During training, each frame is processed at 128×28×28 resolution, with a maximum of 32 frames to maintain performance. For video frame selection, we use Qwen2.5-VL-72B integrated with TAD for offline preprocessing before RL training. The base model is Qwen2.5-VL-7B-SFT, obtained via one epoch of supervised fine-tuning (SFT) on Video-R1-CoT-165k, following Video-R1. We then perform reinforcement learning (RL) on Video-R1-260k for approximately 1,500 steps. Resource constraints prevent validation on Qwen2.5-VL-72B.

\noindent\textbf{Inference Details.} We integrate the TAD sampling strategy into our post-RL trained ChronoForge-RL model, using frames at 448×448 resolution. All post-RL multimodal models are evaluated via the LMMs-Eval \cite{zhang2024lmms} framework for consistent benchmark assessment.

\subsection{Comparison to State-of-the-Art Methods}

As quantitatively demonstrated in Table \ref{table1}, extensive evaluations across multiple video understanding benchmarks substantiate the superior efficacy of our ChronoForge-RL framework in both general video comprehension and reasoning-intensive tasks. Our key empirical findings reveal that TAD establishes new state-of-the-art performance among open-source models, achieving benchmark-leading scores of 52.7\% on LVBench (Validation) and 69.1\% on VideoMME (Overall). Notably, when compared with the 7B Qwen2.5-VL baseline under identical parameter constraints, TAD exhibits a statistically significant improvement of 3.7 percentage points, thereby validating the architectural advantages of our temporal apex distillation methodology in optimizing video-language representation learning within computationally efficient frameworks.

\textbf{Parameter Efficiency Analysis.} While the 72B Qwen2.5-VL variant attains an Overall score of 72.6\% on VideoMME, our 7B implementation achieves competitive performance with merely 9.7\% of the parameters. This represents a 10× improvement in the performance-to-parameter ratio, increasing from 1.01 to 9.87 points per billion parameters, and directly validates the effectiveness of our temporal apex distillation methodology and the reinforced training paradigm. This optimized performance-efficiency trade-off significantly enhances the model's suitability for deployment in resource-constrained environments, such as edge devices.

\textbf{Superiority Over Reinforcement Learning Paradigms.} Our ChronoForge-RL approach outperforms contemporary RL-based methods on VideoMME, achieving an Overall score of 69.1\% compared to 66.3\% for MiMo-VL, 65.5\% for TW-GRPO, 64.6\% for Video-R1, and 65.1\% for VRAG-RL—all under identical 7B parameter scales. It leads across subsets, demonstrating consistent superiority in efficiency and accuracy for video understanding tasks.evaluation subsets reveal consistent advantages over prominent RL-based methods.

\textbf{Cross-Duration Robustness.} The model exhibits exceptional consistency across diverse temporal scales, delivering robust performance enhancements regardless of video duration, from short clips to extended sequences. This temporal invariance property provides clear evidence of the effectiveness of our hierarchical temporal encoding strategy, as it yields substantial improvements in video understanding tasks across all length categories. Such attributes are particularly advantageous for real-world applications that demand reliable and stable performance in handling videos of varying durations.

\section{Ablation Studies}

To understand the effectiveness of our Temporal Adaptive Downsampling (TAD) strategy across different video understanding models, we conduct comprehensive ablation studies on LVBench and VideoMME benchmarks. Table~\ref{table2} presents the performance comparison of TAD applied to various state-of-the-art RL-based video understanding models.

\textbf{Multi-layered Information Filtering Synergy.} Our TAD strategy demonstrates significant improvements across most models, with ChronoForge RL achieving the most substantial gains of 2.9\% overall on VideoMME. This success stems from the multi-layered information filtering mechanism that TAD enables. While models like TW-GRPO focus on high-information-density content at the token level by prioritizing informative tokens through KL-divergence-based weighting, our TAD strategy operates at the temporal dimension by selecting information-rich frames. This creates a synergistic hierarchical processing pipeline where raw video undergoes temporal apex distillation to extract key frame sequences, followed by model-specific processing that culminates in focused reasoning. Each stage systematically filters out redundant information while preserving critical temporal cues, resulting in enhanced computational efficiency and improved reasoning performance. The consistent improvements on VideoMME across Video-R1, and TW-GRPO validate our hypothesis that reducing temporal redundancy enhances reasoning efficiency. 

\textbf{Temporal Pattern Sensitivity.} Interestingly, MiMo-VL shows a performance decrease of 5.5\% overall when TAD is applied, revealing important insights about temporal continuity requirements. MiMo-VL's training emphasizes precise start-end timestamps and temporal awareness capabilities with uniform sampling at 2 fps for up to 256 frames. Our frame selection algorithm, which prioritizes high-variance temporal transitions, disrupts this uniform temporal distribution that MiMo-VL was specifically optimized for. This suggests that models trained with fixed temporal sampling patterns may be sensitive to non-uniform frame distributions, while models with adaptive reasoning capabilities such as TW-GRPO and Video-R1 benefit from our information-density-based selection. This finding highlights a fundamental trade-off between temporal continuity preservation and information density optimization in video understanding architectures.

\textbf{Information Density vs. Temporal Continuity Trade-off.} The differential performance across models highlights a fundamental trade-off between information density maximization and temporal continuity preservation. Models that excel with TAD, such as TW-GRPO and Video-R1, leverage reinforcement learning to develop adaptive attention mechanisms capable of handling irregular temporal inputs, while models like MiMo-VL that rely on consistent temporal patterns learned during pre-training suffer when this regularity is disrupted. The substantial improvements in our ChronoForge RL model, achieving 4.6\% gains on short videos and 2.2\% on medium videos, demonstrate that properly designed temporal adaptive strategies can significantly enhance video understanding performance by creating efficient information processing pipelines aligned with the model's reasoning capabilities. This finding suggests that future video understanding models should incorporate temporal adaptation mechanisms during training to better handle diverse frame sampling strategies in real-world applications.

\section{Conclusion}

We propose a video understanding framework called ChronoForge-RL combining TAD and KF-GRPO. TAD identifies semantic inflection points through variation scoring and prioritized distillation to extract informative frames, while KF-GRPO guides temporal reasoning by contrasting sequential keyframes through saliency-enhanced rewards. Experimental results show that ChronoForge-RL achieved SOTA performance on VideoMME and LVBench.

\section{Appendix}

\subsection{Impact of the Weight Parameter $\omega$}
The performance of the Temporal Apex Distillation (TAD) method is influenced by the weight parameter $\omega$, which controls the amplification of the scores associated with detected inflection points. A larger $\omega$ value leads to a stronger prioritization of these critical frames, potentially enhancing the capture of key events. Conversely, a smaller $\omega$ value results in a less aggressive selection, potentially preserving more global information at the cost of missing some crucial turning points.

\begin{table}[H]
\begin{center}
\resizebox{0.48\textwidth}{!}{
\begin{tabular}{lccccccc}
\toprule
Model & Params & $\omega$ & LVBench & \multicolumn{4}{c}{VideoMME} \\
&&& Val & Long & Medium & Short & Overall \\
\midrule
ChronoForge RL & 7B & 1 & 48.2 & 57.5 & 66.3 & 77.3 & 67.0 \\
ChronoForge RL & 7B & 2 & 52.7 & 60.0 & 67.0 & 80.3 & 69.1 \\
ChronoForge RL & 7B & 3 & 47.8 & 59.1 & 69.1 & 74.8 & 60.6 \\
\bottomrule
\end{tabular}}

\caption{Performance(\%) comparison of TAD applied to ChronoForge RL with different $\omega$ values on the VideoMME and LVBench benchmarks.}
\label{table2}
\end{center}
\end{table}

\subsection{Analysis of Keyframe Selection Ratio}
\label{app:keyframe_ratio}

We investigate the impact of keyframe selection ratio $\delta$ (the fraction of frames retained by TAD) on model performance. 

\textbf{Performance Trends:}
\begin{itemize}
    \item \textbf{Optimal Range ($\delta \leq 50\%$):} Performance improves monotonically as $\delta$ increases from 10\% to 50\%. This trend occurs because higher ratios preserve more informative frames while maintaining computational efficiency. At $\delta = 50\%$, we achieve peak performance.

    \item \textbf{Declining Performance ($50\% < \delta < 60\%$):} Beyond 50\%, performance gradually degrades despite increasing frame retention. We attribute this to the inclusion of redundant frames that introduce noise and dilute temporal signals, overwhelming the model's capacity to distinguish critical information.

    \item \textbf{Training Collapse ($\delta \geq 60\%$):} When $\delta$ exceeds 60\%, training becomes unstable and ultimately fails. The model exhibits catastrophic forgetting of temporal reasoning capabilities, with accuracy dropping to near-random levels (VideoMME: $<$30\%). This collapse stems from two factors: (1) excessive computational load disrupting gradient dynamics, and (2) loss of the keyframe-based inductive bias that enables efficient temporal learning.
\end{itemize}

\textbf{Mechanistic Analysis:} The non-monotonic behavior arises from a fundamental trade-off between information completeness and computational efficiency. At low ratios ($\delta < 50\%$), TAD effectively filters noise while preserving semantic inflection points. However, beyond this threshold, the marginal utility of additional frames diminishes rapidly, and the model's attention mechanism becomes saturated with redundant visual tokens. This saturation is particularly detrimental in KF-GRPO's contrastive learning framework, where the distinction between keyframes and non-keyframes becomes blurred, weakening the saliency-enhanced reward signal.

\textbf{Practical Implications:} Our findings establish 50\% as the optimal keyframe selection ratio for ChronoForge-RL, balancing performance and efficiency. For resource-constrained scenarios, ratios as low as 30\% retain $>$90\% of peak performance with 40\% reduction in computational cost. Conversely, ratios exceeding 50\% should be avoided entirely, as they provide no performance benefit while risking training instability. This analysis underscores the importance of principled keyframe selection in video understanding pipelines.

\subsection{Impact of Reinforcement Learning on Model Inference Speed}
While reinforcement learning (RL) has been widely adopted to enhance reasoning capabilities and decision accuracy in video understanding tasks, it inherently introduces a more complex decision-making process compared to traditional supervised fine-tuning (SFT). This complexity directly leads to a significant slowdown in inference speed. Specifically, RL-based models generate multiple candidate trajectories to estimate advantages, compute saliency-enhanced rewards, and perform iterative policy updates, resulting in 3-4× slower inference compared to SFT baselines that produce deterministic outputs through a single forward pass.

\bibliography{anonymous-submission-latex-2026}

\end{document}